%% file: paper.tex
\documentclass[lettersize,journal]{IEEEtran}

\usepackage[utf8]{inputenc}
\usepackage{csquotes}

\usepackage{amsmath,amssymb,amsfonts}
\usepackage{siunitx}
\usepackage{graphicx}
\usepackage{tikz}

\usepackage{array}
\usepackage{makecell}
\usepackage{longtable}
\usepackage{xcolor}
\usepackage{listings}
\usepackage{stfloats}
\usepackage{textcomp}
\usepackage{url}
\usepackage{comment}

\usepackage{acro}
\input{abkuerzungen}

\usepackage[backend=biber,style=ieee]{biblatex}
\begin{document}

\title{Implementation and evaluation of a prediction algorithm for an autonomous vehicle}
 
\author{
    \textbf{Marco Leon Rapp}\\
    marcoleon.rapp@gmail.com
}

\maketitle

\section*{Abstract}
\textbf{
This paper presents a prediction algorithm that estimates the vehicle trajectory every five milliseconds for an autonomous vehicle. A kinematic and a dynamic bicycle model are compared, with the dynamic model exhibiting superior accuracy at higher speeds. Vehicle parameters such as mass, center of gravity, moment of inertia, and cornering stiffness are determined experimentally. For cornering stiffness, a novel measurement procedure using optical position tracking is introduced.
The model is incorporated into an extended Kalman filter and implemented in a ROS node in C++. The algorithm achieves a positional deviation of only 1.25 cm per meter over the entire test drive and is up to 82.6\% more precise than the kinematic model.
}

\section{Introduction}
\IEEEPARstart{T}{he} CAuDri Challenge is a student competition in which autonomous scale vehicles (1:10) compete against each other in various disciplines. To participate, the Smart Rollerz team from DHBW Stuttgart develops the vehicle \textit{Smarty}, which uses a camera to perceive its surroundings. The trajectory is controlled via lateral and longitudinal controllers, which actuate both the electric motor and the steering.

New information about the vehicle pose can only be acquired from the camera every 35 milliseconds. To enable control at intervals of five milliseconds, a prediction algorithm is proposed to estimate the vehicle's current position and orientation in the gaps between camera measurements. A speed sensor on the rear axle measures the vehicle's velocity, the current steering angle is known from the control system, and an inertial measurement unit (IMU) could additionally record accelerations and angular rates to further improve the prediction.

\begin{figure}[!t]
  \centering
  \includegraphics[width=3.5in]{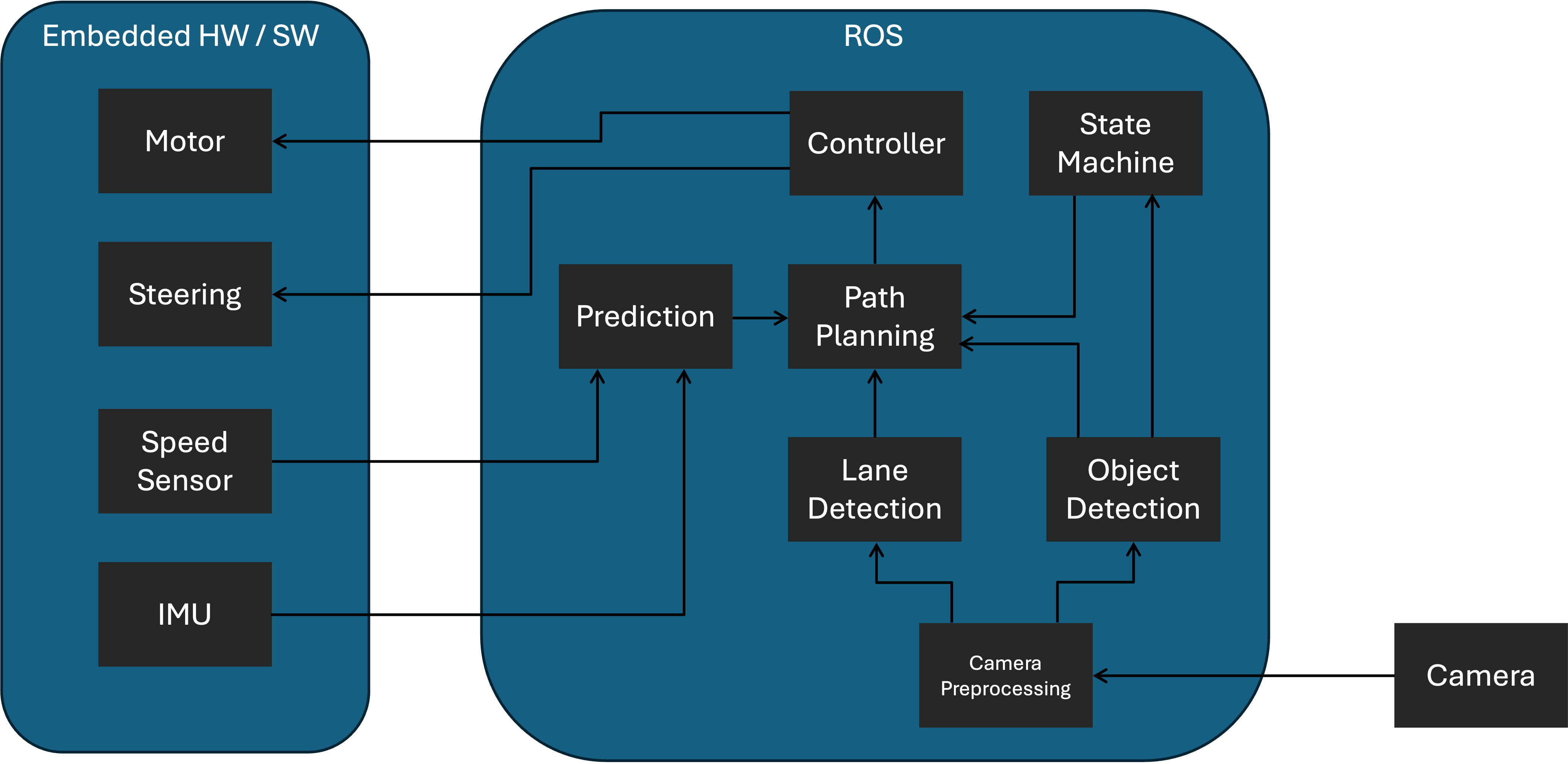}
  \caption{Architecture of the ROS Nodes}
  \label{fig:Softwarestruktur}
  \end{figure}

For the prediction algorithm to be reliably implemented, vehicle models must be analyzed and tested, and vehicle-specific parameters must be determined. In the absence of a test bench, these parameters cannot be directly measured, and must therefore be acquired experimentally or calculated from known quantities. The objective is to implement the best-performing model in a ROS node, which processes all sensor data, predicts the vehicle pose, and provides the estimated values to the control system. Figure~\ref{fig:Softwarestruktur} illustrates the architecture of the ROS2 nodes onboard the vehicle.

\section{Fundamentals}
The prediction model uses a bicycle model to predict how the vehicle's position and orientation change by integrating angular velocity and speeds. The dynamic bicycle model additionally allows consideration of the lateral velocity, which, at higher speeds with non-zero slip, is not negligible \cite{Zomotor1991}.

\subsection{Kinematic Bicycle Model}

\noindent This model describes the vehicle motion solely through geometric relationships without considering dynamic forces. It predicts the trajectory reasonably well for low speeds and small time intervals.

\noindent \textbf{State variables \cite{7225830}:}
\begin{itemize}
    \item $X, Y$: position of the vehicle's center of gravity
    \item $\psi$: yaw angle
    \item $v_{\mathrm{x}}$: constant longitudinal velocity
\end{itemize}
\vspace{0.3cm}

\noindent \textbf{Model assumptions:}
\begin{itemize}
    \item No rear-axle steering: $\delta_h \approx 0$ \cite{7225830}
    \item The effective steering angle is $\delta_v$ \cite{Ammon1997}.
\end{itemize}
\vspace{0.3cm}

\noindent \textbf{Model equations: \cite{7225830}}
\begin{align}
    \dot{\psi} &= \frac{v_{\mathrm{x}}}{\ell_{\mathrm{v}} + \ell_{\mathrm{h}}} \tan(\delta_{\mathrm{v}}) \\
    \dot{X} &= v_{\mathrm{x}} \cos(\psi) \\
    \dot{Y} &= v_{\mathrm{x}} \sin(\psi)
\end{align}

\subsection{Dynamic Bicycle Model}

\begin{figure}[!t]
  \centering
  \includegraphics[width=3.5in]{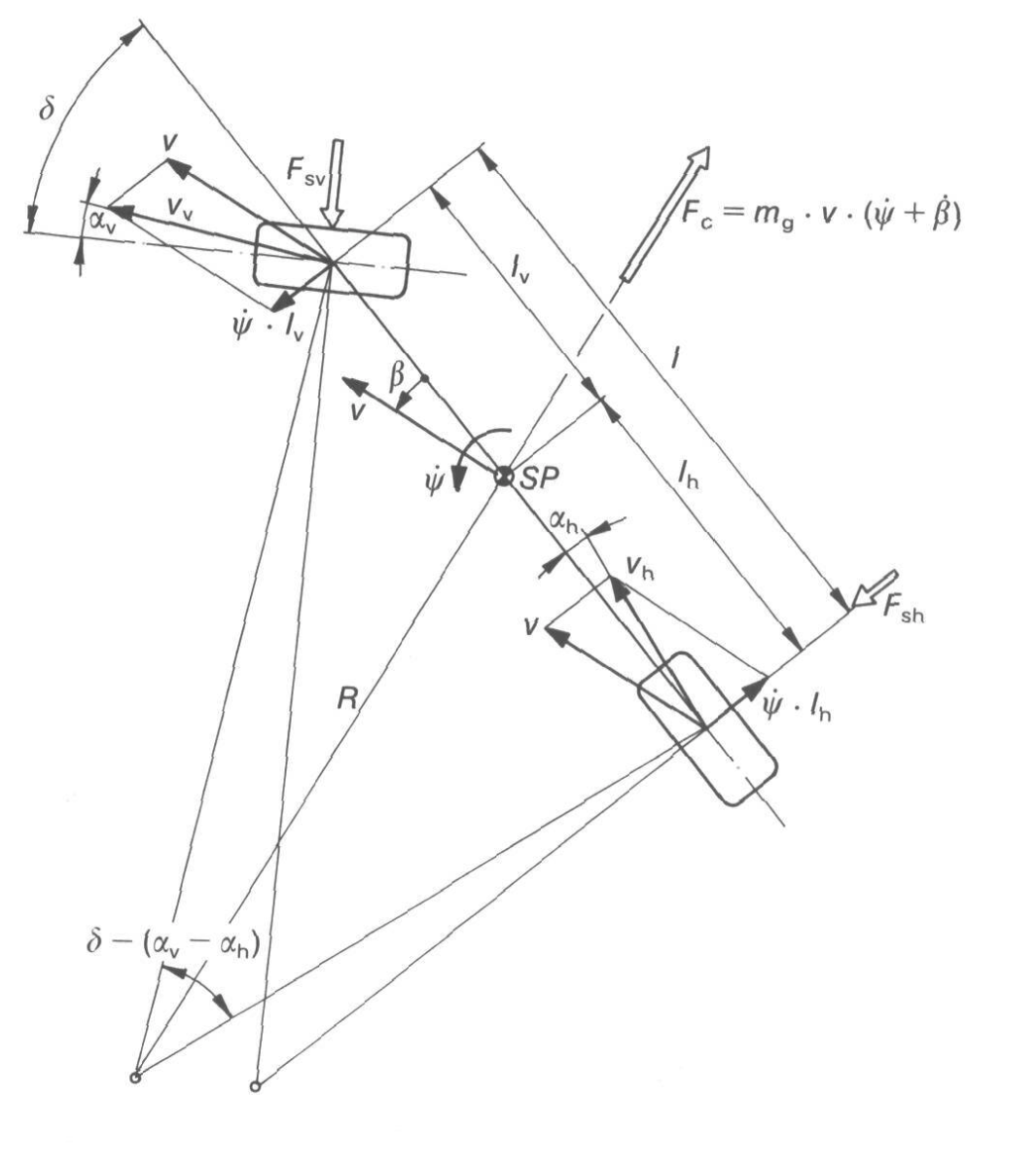}
  \caption{Dynamic bicycle model \cite{Zomotor1991}}
  \label{fig:DynamischesModell}
\end{figure}

\noindent The dynamic model extends the kinematic model by incorporating lateral forces, inertia, slip angle, and drift angle \cite{Riekert1940, Zomotor1991}. This allows it to predict more accurately at higher speeds and during dynamic maneuvers. Figure~\ref{fig:DynamischesModell} shows the dynamic bicycle model.
\vspace{0.3cm}

\noindent \textbf{Additional state variables \cite{Zomotor1991}:}
\begin{itemize}
    \item $v_y$: lateral velocity
    \item $\beta$: drift angle
    \item $\dot{\psi}, \ddot{\psi}$: yaw rate and yaw acceleration
\end{itemize}
\vspace{0.3cm}
\noindent \textbf{Forces \cite{Zomotor1991}:}
\begin{align}
    F_{\mathrm{sv}} &= C_{\mathrm{v}} \cdot \alpha_{\mathrm{v}} \\
    F_{\mathrm{sh}} &= C_{\mathrm{h}} \cdot \alpha_{\mathrm{h}} \label{eq:Fsh}
\end{align}
\vspace{0.3cm}
\noindent \textbf{Slip angles \cite{Zomotor1991}:}
\begin{align}
    \alpha_\mathrm{v} &= \delta - \beta - \frac{\dot{\psi} \ell_\mathrm{v}}{v} \label{eq:alpha_v_alternative} \\
    \alpha_\mathrm{h} &= -\beta + \frac{\dot{\psi} \ell_\mathrm{h}}{v} 
\end{align}
\textbf{State equations \cite{Zomotor1991, 9669260}:}
\begin{align}
    \dot{X} &= v_x \cos \psi - v_y \sin \psi \label{eq:x_dot} \\
    \dot{Y} &= v_y \cos \psi + v_x \sin \psi \\
    \dot{v}_x &= a_x + v_y \dot{\psi} - \frac{F_{\mathrm{sv}} \sin \delta}{m} \\
    \dot{v}_y &= -v_x \dot{\psi} + \frac{F_{\mathrm{sv}} \cos \delta + F_{\mathrm{sh}}}{m} \\
    \ddot{\psi} &= \frac{\ell_{\mathrm{v}} F_{\mathrm{sv}} \cos \delta - \ell_{\mathrm{h}} F_{\mathrm{sh}}}{J_z} \label{eq:psidot_dot}
\end{align}

\subsection{Extended Kalman Filter}

The \acs{ekf} is a filter used to estimate the state of a dynamic system based on measurements that are nonlinearly related to the system state \cite{Wendel+2011}. In \cite{zhang2023multisensor}, it is shown how the \ac{ekf} can be applied with a kinematic bicycle model for accurate position estimation in a \ac{slam} system.

\noindent The system is described by its state equation as a function of the previous state and the system inputs \cite{Wendel+2011}:
\begin{equation}
    x_{k+1} = f(x_{k}, u_{k}) + \mathbf{G}_{k} \, w_{k},
    \label{eq:system_model}
\end{equation}

\noindent $x_{k+1}$ is the discrete-time state at step $k+1$. It is calculated using the nonlinear, discrete-time system $f(x_{k}, u_{k})$. $u_k$ is the input vector \cite{Wendel+2011}.

\noindent The measurement function $h_k$ describes the nonlinear relationship between the measurement vector and the state vector \cite{Wendel+2011}:
\begin{equation}
    \widetilde{y}_k = h_k(x_k) + v_k
    \label{eq:measurement_model}
\end{equation}
The measurement noise $v_k$ is assumed to be zero-mean and normally distributed.

\noindent The functions ($f, h$) are linearized through the following Jacobian matrices \cite{Wendel+2011}:

\vspace{0.3cm}
\noindent \textbf{Transition matrix} $\mathbf{\Phi}_k$ (Jacobian matrix of $f$):
\begin{equation}
    \mathbf{\Phi}_k = \left. \frac{\partial f(x_k, u_k)}{\partial x} \right|_{x=\hat{x}_k}
    \label{eq:transition_matrix}
\end{equation}

\noindent \textbf{Measurement matrix} $\mathbf{H}_k$ (Jacobian matrix of $h$):
\begin{equation}
    \mathbf{H}_k = \left.\frac{\partial h_k(x_k)}{\partial x_k}\right|_{x_k=\hat{x}_k}
    \label{eq:measurement_matrix}
\end{equation}

\vspace{0.3cm}
The Jacobian matrices can be computed analytically from the functions ($f, h$). If this is not possible, they can be approximated numerically online. However, this introduces additional computational effort into the estimation, which makes the algorithm less efficient \cite{marchthaler2024nichtlineare}. For this reason, the matrices are computed offline in this work.

In the \ac{ekf} algorithm, the system state $x_k$ is first predicted using the system model and then corrected with measurements. $\hat{x}_k$ denotes the estimated state. The prediction estimate is referred to as the \textit{a priori} estimate, which is denoted in the following as $\hat{x}_{k+1}^-$. The corrected estimate is referred to as the \textit{a posteriori} estimate $\hat{x}_k^+$ \cite{Wendel+2011}:
\vspace{0.3cm}
\begin{enumerate}
    \item \textbf{Prediction}:
    \begin{align}
        \hat{x}_{k+1}^- &= f(\hat{x}_k^+, u_k), \label{eq:state_prediction} \\
        \mathbf{P}_{k+1}^- &= \mathbf{\Phi}_k \mathbf{P}_k^+ \mathbf{\Phi}_k^\top + \mathbf{Q}_k \label{eq:covariance_prediction}
    \end{align}
    
    \item \textbf{Correction}:
    \begin{align}
        \mathbf{K}_k &= \mathbf{P}_k^- \mathbf{H}_k^\top \left( \mathbf{H}_k \mathbf{P}_k^- \mathbf{H}_k^\top + \mathbf{R}_k \right)^{-1} \label{eq:kalman_gain} \\
        \hat{x}_k^+ &= \hat{x}_k^- + \mathbf{K}_k \left( y_k - h_k(\hat{x}_k^-) \right) \label{eq:state_update} \\
        \mathbf{P}_k^+ &= \left( \mathbf{I} - \mathbf{K}_k \mathbf{H}_k \right) \mathbf{P}_k^- \label{eq:covariance_update}
    \end{align}
\end{enumerate}

The \textit{a priori} estimate is predicted in the first part of the algorithm using the system model \eqref{eq:system_model}. Together with a sensor measurement, this can be used in the second part to compute the \textit{a posteriori} estimate, where the weighting of the values can be specified by the covariance matrices $\mathbf{Q}_k$ and $\mathbf{R}_k$, and is combined in the Kalman gain $\mathbf{K}_k$. After each estimation step, the estimation uncertainty $\mathbf{P}$ is calculated, which also indicates the quality of the estimate. $\mathbf{R}_k$ represents the measurement noise, while $\mathbf{Q}_k$ represents the system inaccuracies \cite{Wendel+2011}. \\
The identity matrix is denoted here as $\mathbf{I}$.

\section{Parameter Determination}

In \cite{7225830}, a kinematic bicycle model serves as the basis for a \ac{mpc} controller. This reduces computational load, which is particularly relevant when many prediction steps are involved. In this work, one step per iteration suffices.  
For slow transport platforms such as in \cite{simon_2023}, the kinematic model is sufficient despite physical limitations, as slip is negligible.  
A nonlinear \ac{mpc} in \cite{pei_2025} also uses a kinematic model, since the vehicle does not exceed \SI{10}{\kilo\meter\per\hour} during parking maneuvers.  
Similarly, \cite{xuehao_2022} employs the kinematic model for trajectory planning in a robot with axle steering. Due to low speeds ($< \SI{1}{\meter\per\second}$) and real-time requirements, slip effects are negligible.  
The dynamic model in \cite{xing_2023} is used to replicate human driving behavior. It achieves sufficient accuracy with moderate computational effort.

For the prediction algorithm in the autonomous vehicle, a dynamic bicycle model is used because computational effort is not a limiting factor. Instead, the prediction should be as accurate as possible to minimize long-term error. \\
For this purpose, the following parameters must be determined:

\begin{enumerate}
    \item Total mass
    \item Distance between the vehicle's center of gravity and the front and rear axles
    \item Moment of inertia about the vertical axis
    \item Cornering stiffnesses of the front and rear axles
\end{enumerate}

\noindent Measurement methods from the literature cannot be fully adopted, as the vehicle is not a full size passenger car. The following measurement techniques are considered to be most suitable for the autonomous vehicle. \\
The total mass of the vehicle can be measured using a scale. It is important that all components are attached to the vehicle, such as the battery.

\subsection{Vehicle Center of Gravity}

\begin{figure}[!t]
    \centering
    \includegraphics[width=3.0in]{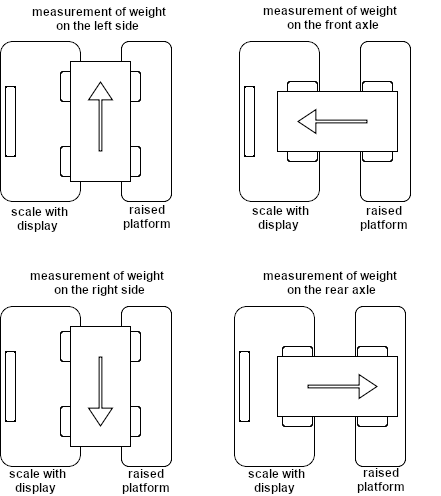}
    \caption{Schematic setup of the center-of-gravity measurement}
    \label{fig:schwerpunkt_schema}
\end{figure}

The vehicle's center of gravity characterizes the average position of the mass distribution \cite{8373334}. \\
By rearranging the equations from \cite{8047738} with respect to $\ell_{\mathrm{h}}$ and $\ell_{\mathrm{v}}$, one obtains the distances between the center of gravity and the front or rear axle, respectively. The total mass, wheelbase, and the forces acting between the axles and the road surface must be known. The gravitational acceleration constant $g$ is assumed to be approximately $9.81$:

\begin{equation}
    \ell_{\mathrm{h}} = \frac{F_\mathrm{gv} \ell}{m g}
    \hspace{2.0cm}
    \ell_{\mathrm{v}} = \frac{F_\mathrm{gh} \ell}{m g}
\end{equation}
The forces $F_\mathrm{gv}$ and $F_\mathrm{gh}$ are the forces acting perpendicular to the front and rear axles, respectively. Similarly, forces on the left and right sides of the vehicle can also be determined. Figure~\ref{fig:schwerpunkt_schema} illustrates how the vehicle was placed on a scale to measure the four distances to the center of gravity. The raised platform is equal in height to the scale, ensuring that the vehicle remains level.

\subsection{Moment of Inertia}

\begin{figure}[!t]
    \centering
    \includegraphics[width=3.0in]{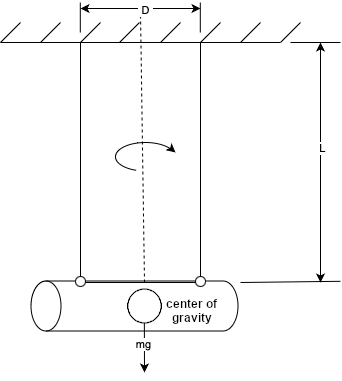}
    \caption{Schematic setup of the bifilar pendulum with a suspended object}
    \label{fig:bifilar_schema}
\end{figure}

In \cite{8049885, hamid_2023}, a bifilar pendulum is used to determine the moments of inertia of quadcopters. Figure~\ref{fig:bifilar_schema} shows the schematic setup of the pendulum. The quadcopter is suspended by two cords of length $L$, spaced a distance $D$ apart, with the quadcopter's center of gravity positioned midway between them. A gravitational force $F_g$ acts on the quadcopter. \\
The moment of inertia $J$ can then be calculated according to \cite{hamid_2023}:

\begin{equation}
J = \frac{mgD^2T^2}{16\pi L}
\label{eq:traegheitsmoment}
\end{equation}

\noindent The moment of inertia of the autonomous model car can also be determined using this method. The vehicle is suspended as described above and rotated about the vertical axis. This rotation induces torsional forces, causing the vehicle to oscillate \cite{hamid_2023}. By averaging several oscillation periods, the mean period $T$ is obtained and substituted into equation~\ref{eq:traegheitsmoment}.

\subsection{Cornering Stiffnesses}

\begin{figure}[!t]
    \centering
    \includegraphics[width=3.5in]{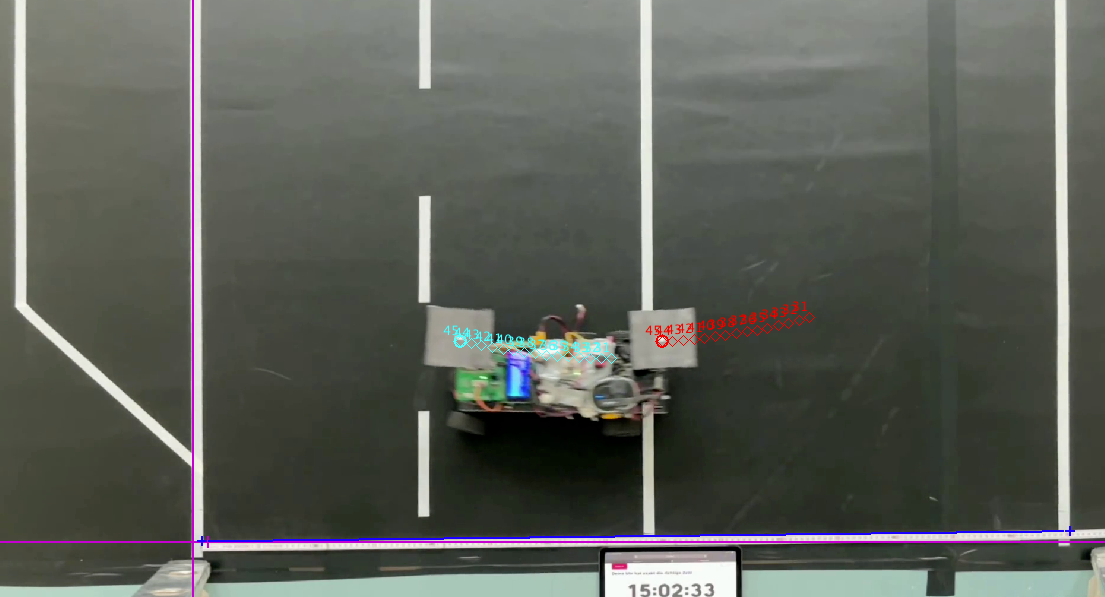}
    \caption{Screenshot from the tracking software}
    \label{fig:MotionTracking}
\end{figure}

Cornering stiffness describes the relationship between the slip angle and the lateral force acting on the wheel, and depends on both the vehicle and the road surface. Although it is not possible to measure the cornering stiffnesses directly, there are methods for online identification \cite{Zomotor2002, 9204763}. Since the autonomous vehicle always operates on the same surface, online estimation is not required. \\
The following presents a new approach for calculating the cornering stiffnesses using optical tracking:\\
By rearranging \eqref{eq:Fsh} and equating it with \eqref{eq:alpha_v_alternative}, we obtain
\begin{equation}
C_\mathrm{v} = \frac{F_\mathrm{sv}}{\delta - \beta - \frac{\dot{\psi} \ell_{\mathrm{v}}}{v}}
\label{eq:cornering_stiffness_front}
\end{equation}
with
\begin{equation}
    F_\mathrm{sv} = \frac{\ell_{\mathrm{h}}}{\ell} m \left( a_y \cos(\delta - \alpha_\mathrm{v}) + a_x \sin(\delta - \alpha_\mathrm{v}) \right)
    \label{eq:querkraft_vorne}
\end{equation}
where $a_x \approx 0$ is assumed, and the equation is simplified for small angles to
\begin{equation}
F_\mathrm{sv} = \frac{\ell_{\mathrm{h}}}{\ell} m a_y
\label{eq:querkraft_vorne_approx}
\end{equation}
Similarly, the following relationship for the rear axle can be derived:
\begin{equation}
C_\mathrm{h} = \frac{F_\mathrm{sh}}{- \beta + \frac{\dot{\psi} \ell_{\mathrm{h}}}{v}}, \quad
F_\mathrm{sh} = \frac{\ell_{\mathrm{v}}}{\ell} m a_y
\label{eq:cornering_stiffness_rear}
\end{equation}
The vehicle mass and the lengths used above are known. The lateral acceleration $a_y$ can be measured directly using an \ac{imu} mounted at the vehicle's center of gravity. The \ac{imu} also measures the yaw rate $\dot{\psi}$. The velocity is either measured by a speed sensor during the experiment or is obtained optically from the change in the center-of-gravity position over time. \\
An optical measurement method is used to determine the drift angle $\beta$, since this angle cannot be measured directly. The drift angle is defined as the angle between the tangent to the trajectory and the vehicle's heading, and it occurs during cornering. In Figure~\ref{fig:DynamischesModell}, it is shown as the angle between the velocity vector and the vehicle's longitudinal axis. \\
To calculate it, the vehicle performs a steady-state circular drive at constant speed and constant steering angle. The vehicle is filmed from above with a camera. Markers are attached to the front and rear of the vehicle, and their positions can be tracked in the video using software. \\
Figure~\ref{fig:MotionTracking} shows how the markers are tracked in the software to determine their positions. To compute $\beta$, the position of the center of gravity must be calculated. From the sequence of center-of-gravity positions, tangents can be computed for each moment in time, which, together with the vehicle heading, define the drift angle.

By substituting the measured values into the equations derived above, the cornering stiffnesses can be calculated. It should be noted that the accuracy of the results can be significantly affected by errors if the camera resolution or frame rate is too low, or if image distortion from the lens or camera perspective is excessive. Furthermore, slip angles large enough to compute the cornering stiffnesses occur only at sufficiently high speeds. At excessively high speeds, however, the tire characteristic curve leaves the approximately linear range \cite{Ammon1997}. For this reason, the measurements shown above were performed at speeds of around one meter per second.

\section{Implementation}

The \ac{ros} node is intended to predict the trajectory at 200~Hz. For this purpose, the continuous-time equations \eqref{eq:x_dot}--\eqref{eq:psidot_dot} are not suitable. In \cite{9669260}, the equations were formulated in discrete-time form:

\begin{flalign}
X_k &= X_{k-1} + \Delta t \cdot \left( v_{x,k-1} \cos \psi_{k-1} - v_{y,k-1} \sin \psi_{k-1} \right) &&  \label{eq:discrete_X} \\
Y_k &= Y_{k-1} + \Delta t \cdot \left( v_{x,k-1} \sin \psi_{k-1} + v_{y,k-1} \cos \psi_{k-1} \right) && \label{eq:discrete_Y} \\
\psi_k &= \psi_{k-1} + \Delta t \cdot \dot{\psi}_{k-1} && \label{eq:discrete_psi} \\
v_{x,k} &= v_{x,k-1} + \Delta t a_{x,k} && \label{eq:discrete_vx} \\
v_{y,k} &= \frac{a - b}{m v_{x,k-1} - \Delta t \left( C_\mathrm{v} + C_\mathrm{h} \right)} && \label{eq:discrete_vy} \\
\dot{\psi}_k &= \frac{c - d}{J_z v_{x,k-1} - \Delta t \left( \ell_{\mathrm{v}}^2 C_\mathrm{v} + \ell_{\mathrm{h}}^2 C_\mathrm{h} \right)} && \label{eq:discrete_psidot} 
\end{flalign}
\noindent where
\begin{align*}
a &= m v_{x,k-1} v_{y,k-1} + \Delta t \left( \ell_{\mathrm{v}} C_\mathrm{v} - \ell_{\mathrm{h}} C_\mathrm{h} \right) \dot{\psi}_{k-1} \\
b &= \Delta t C_\mathrm{v} \delta_k v_{x,k-1} - \Delta t m v_{x,k-1}^2 \dot{\psi}_{k-1} \\
c &= J_z v_{x,k-1} \dot{\psi}_{k-1} + \Delta t \left( \ell_{\mathrm{v}} C_\mathrm{v} - \ell_{\mathrm{h}} C_\mathrm{h} \right) v_{y,k-1} \\
d &= \Delta t \ell_{\mathrm{v}} C_\mathrm{v} \delta_k v_{x,k-1}
\end{align*}

\subsection{Extended Kalman Filter}

The discrete-time model predicts the state. The longitudinal velocity $v_x$ and yaw rate $\dot{\psi}$ can also be measured directly using a speed sensor and \ac{imu}, respectively. Since the model is nonlinear, an \ac{ekf} is suitable for fusing the prediction with sensor data. \\
The measured longitudinal acceleration from the \ac{imu} and the steering angle at the front axle are system inputs in the model \eqref{eq:discrete_vx}--\eqref{eq:discrete_psidot}. Therefore, the system input vector is:

\begin{equation} 
    u_k = \begin{bmatrix}
        \delta_k \\
        a_{x,k} 
    \end{bmatrix} 
    \label{eq:input_vector_u} 
\end{equation}
The measurement vector is:
\begin{equation}
    z = \begin{bmatrix}
        \dot{\psi} \\
        v_x
    \end{bmatrix}
    \label{eq:spaltenvektor}
\end{equation}
containing the yaw rate and the longitudinal velocity. Since $\dot{\psi}$ and $v_x$ are state variables, the measurement equation is:

\begin{equation}
    h(x) = \begin{bmatrix}
        \dot{\psi} \\
        v_x
    \end{bmatrix}
    \label{eq:spaltenvektor}
\end{equation}

The covariance matrices $\mathbf{Q}$ and $\mathbf{R}$ describe the uncertainties of the system model and the measurement noise respectively, using variances and their correlations. For the \ac{imu}, the variances can be calculated from measurement data obtained during the stationary circular driving experiment. \\
The variance of the longitudinal acceleration is part of the system uncertainty and cannot be directly inserted into the matrices. Since the uncertainty of the longitudinal velocity includes the variance of the acceleration sensor, the variance for $v_x$ must therefore satisfy $\sigma_{v_x}^2 > \Delta t\, \sigma_{a_x}^2$. The measurement noise for $\dot{\psi}$ can be directly entered into the $\mathbf{R}$ matrix, but it should be multiplied by a factor of ten to compensate for additional uncertainties and vibrations.

Since $\dot{\psi}$ in the prediction step is highly dependent on the accuracy of lateral velocity estimation and therefore subject to certain modeling inaccuracies, direct measurement of this quantity is suitable for compensating these uncertainties. This allows not only errors in determining the cornering stiffnesses to be corrected, but also errors that occur, for example, when the road surface changes. For this purpose, the weighting of the measurement should be set sufficiently high. The following covariance matrices have led to very good results:

\[
\mathbf{Q} = 
\begin{bmatrix}
\sigma_{v_x}^2 & \mathrm{Cov}(v_x, v_y) & \mathrm{Cov}(v_x, \dot{\psi}) \\
\mathrm{Cov}(v_y, v_x) & \sigma_{v_y}^2 & \mathrm{Cov}(v_y, \dot{\psi}) \\
\mathrm{Cov}(\dot{\psi}, v_x) & \mathrm{Cov}(\dot{\psi}, v_y) & \sigma_{\dot{\psi}}^2 
\end{bmatrix}
\]

\[
=
\begin{bmatrix}
0.05 & 0 & 0 \\
0 & 0.01 & 0 \\
0 & 0 & 0.01
\end{bmatrix}
\]

\noindent The model uncertainties of equations \eqref{eq:discrete_vx}--\eqref{eq:discrete_psidot} are entered on the main diagonal.

\[
\mathbf{R} = 
\begin{bmatrix}
\sigma_{\dot{\psi}}^2 & \mathrm{Cov}(\dot{\psi}, v_x) \\
\mathrm{Cov}(v_x, \dot{\psi}) & \sigma_{v_x}^2 \\
\end{bmatrix}
=
\begin{bmatrix}
0.125 & 0 \\
0 & 0.10 \\
\end{bmatrix}
\]

\noindent The variance $\sigma_{\dot{\psi}}^2$ is based on the measured variance of the sensor. $\sigma_{v_x}^2$ was estimated due to the lack of suitable measurements. \\
The \ac{ekf} is implemented in MATLAB, from which efficient C++ code can be generated. This enables the predictions to be calculated efficiently in a \ac{ros} node.
\section{Validation}

\begin{figure}[!t]
    \centering
    \includegraphics[width=3.5in]{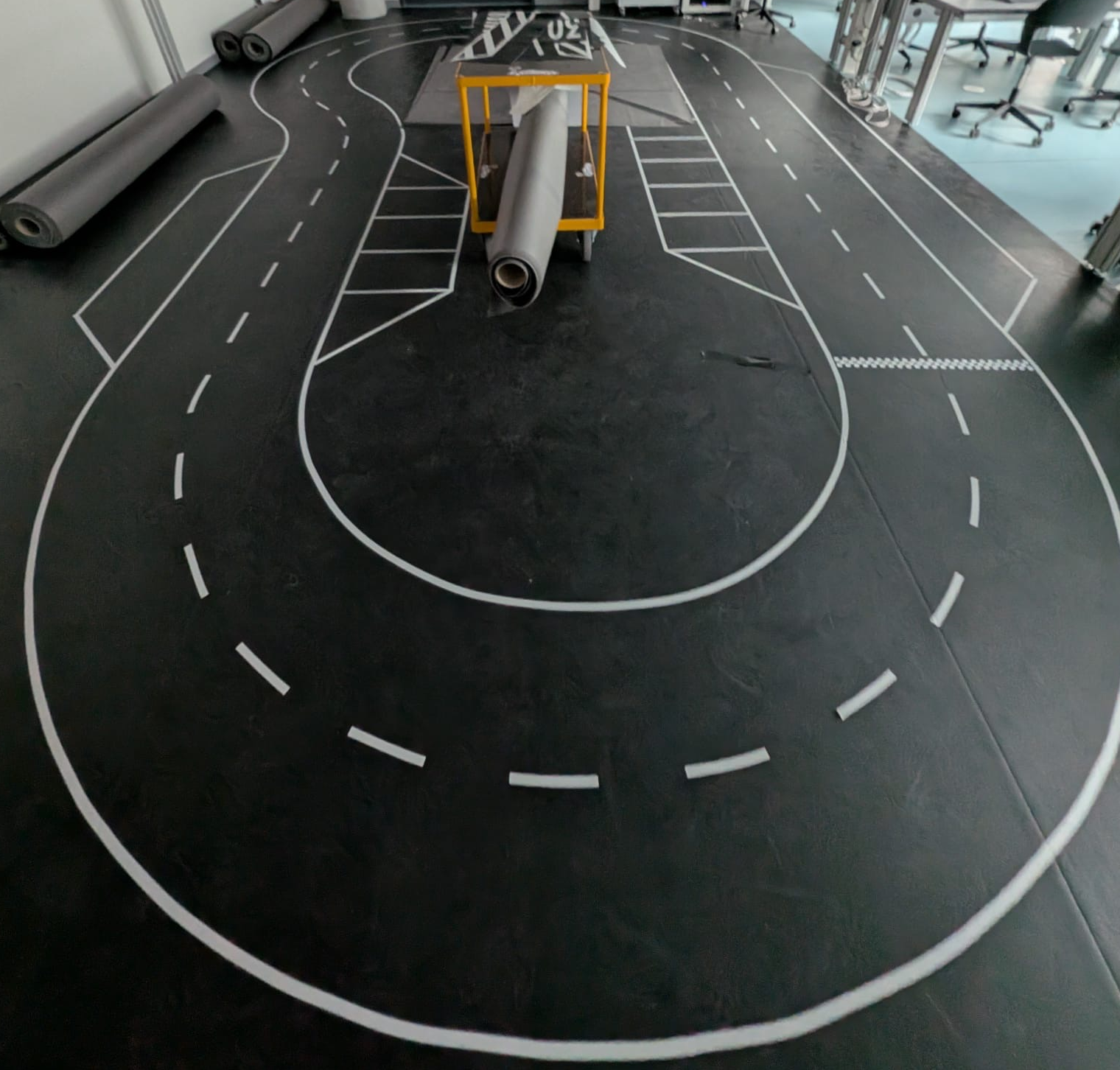}
    \caption{Test track in the Smart Rollerz laboratory}
    \label{fig:Teststrecke}
\end{figure}

To assess how accurately the sensor fusion model performs with the measured parameters, a single lap is driven on the test track. Figure~\ref{fig:Teststrecke} shows the test track. The start point is located at the bottom right; the vehicle begins driving counterclockwise.

\begin{figure}[!t]
    \centering
    \includegraphics[width=3.5in, trim=19cm 0 19cm 0, clip]{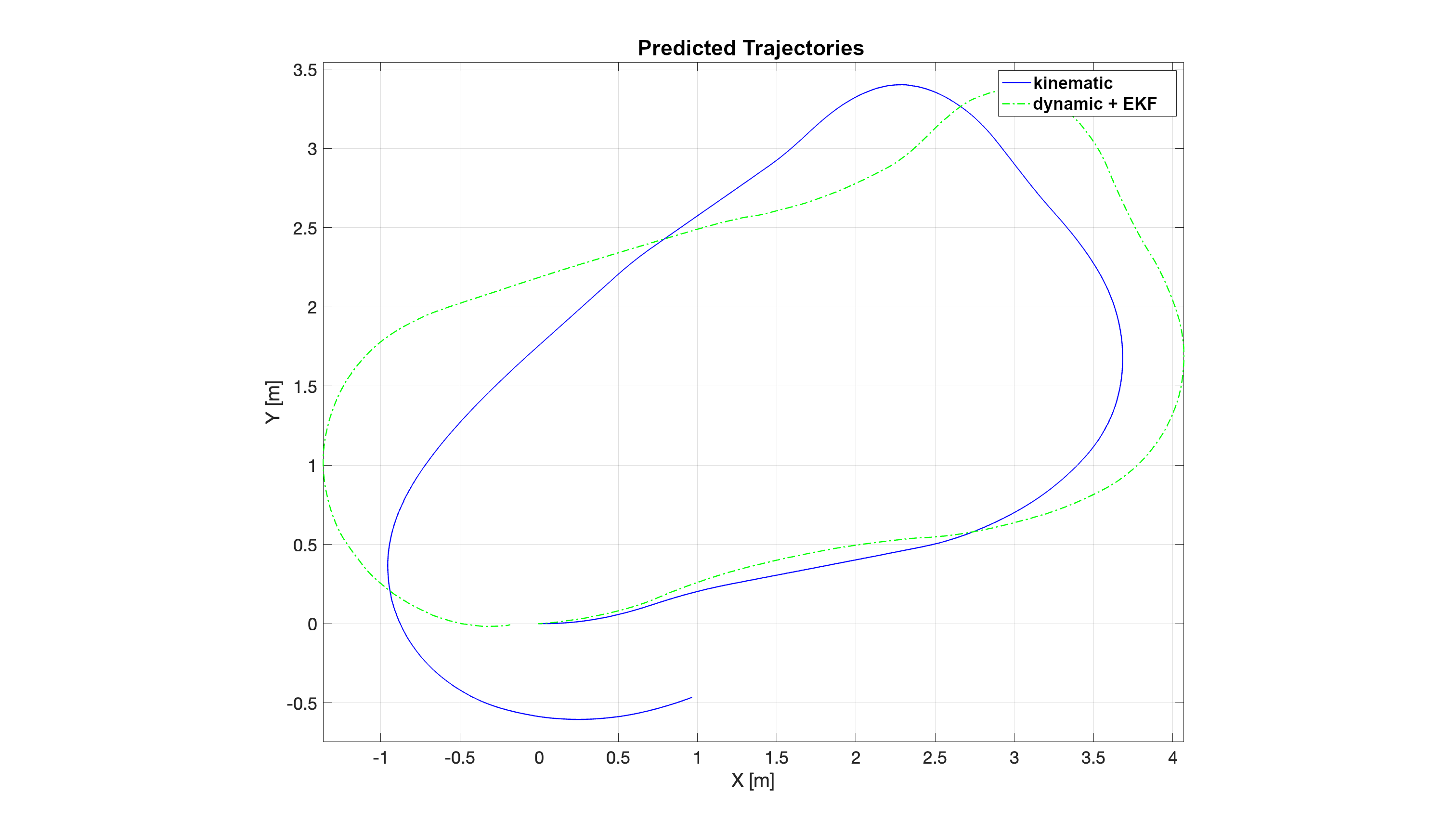}
    \caption{Comparison of predicted trajectories}
    \label{fig:Trajektorie_Alt2}
\end{figure}

\noindent The accuracy of the prediction is expressed as the deviation of the estimated position and orientation from the true value. To compute this error for every prediction step, the reference values would have to be measured with a more accurate measurement system. Since such a system is not available, the vehicle is guided to return as precisely as possible to the starting point at the end of the lap. The final predicted values can therefore be compared with the initial state to approximately determine the deviation from the true state. \\
Figure~\ref{fig:Trajektorie_Alt2} shows the predicted trajectory. As a reference, the prediction of the dynamic model with \ac{ekf} (green) is compared to the prediction of a kinematic model (blue). The start point is $(0,0)$.

As can be seen, the endpoint of the proposed model lies much closer to the starting point than that of the kinematic model. Furthermore, the kinematic model deviates far more strongly in curves from the test track's shape visible in Figure~\ref{fig:Teststrecke}.  
Comparing the average deviation per meter traveled, the proposed model, with a deviation of only 1.25~cm per meter, is up to 82.6\% more accurate than the kinematic model. The deviation in final orientation compared to the initial yaw angle is 2.9$^{\circ}$ for the dynamic model, versus 18.6$^{\circ}$ for the kinematic model. \\
In the intended application described initially, however, long-term deviation is of lesser importance. Since only 35 milliseconds must be bridged until the camera acquires a new image, and predictions are made every five milliseconds, the key metric is the average deviation after seven prediction steps. This value is less than one millimeter, indicating that the proposed model is highly suitable for the application. In the event of delays in receiving image data from the camera, the model is capable of continuing to predict position and orientation with minimal error.

\section{Summary}

In this work, a prediction model for an autonomous vehicle was presented, which is 82.6\% more accurate over the long term than a kinematic model. \\
For this purpose, the vehicle parameters required for a dynamic bicycle model were measured. The total mass as well as the longitudinal and lateral center of gravity could be determined using a scale. For the moment of inertia, a measurement method from the field of aviation was adapted, in which quadcopters are suspended from a bifilar pendulum and the moment of inertia about the rotation axis is determined from the oscillation period. \\
Measuring cornering stiffnesses without a test bench posed a particular challenge in vehicle dynamics modeling. To address this, a new method was introduced in which the quantities in the formulas were determined using optical tracking and an \ac{imu}. This approach can be implemented without a test bench or similar equipment, but is affected by inaccuracies in optical tracking. \\
By fusing the prediction model with sensor data, the longitudinal velocity and yaw rate can be corrected with measurements, thereby compensating for system inaccuracies and making the model highly accurate.

\printbibliography

\end{document}

%% file: abkuerzungen.tex
\DeclareAcronym{mpc}{
  short = MPC,
  long = Model Predictive Control
}

\DeclareAcronym{imu}{
  short = IMU,
  long = Inertial Measurement Unit
}

\DeclareAcronym{caudri}{
  short = CAuDri,
  long = Cognitive Autonomous Driving
}

\DeclareAcronym{nuc}{
  short = NUC,
  long = Next Unit of Computing
}

\DeclareAcronym{ros}{
  short = ROS,
  long = Robotic Operating System
}

\DeclareAcronym{mems}{
  short = MEMS,
  long = mikroelektromechanisches System
}

\DeclareAcronym{aruco}{
  short = ArUco,
  long = Augmented Reality University of Cordoba
}

\DeclareAcronym{dof}{
  short = DOF,
  long = Degrees Of Freedom
}

\DeclareAcronym{uart}{
  short = UART,
  long = Universal Asynchronous Receiver Transmitter
}

\DeclareAcronym{i2c}{
  short = I$^2$C,
  long = Inter-Integrated Circuit
}

\DeclareAcronym{cad}{
  short = CAD,
  long = Computer Aided Design
}

\DeclareAcronym{ekf}{
  short = EKF,
  long = Erweitertes Kalman-Filter
}

\DeclareAcronym{slam}{
  short = SLAM,
  long = Simultaneous Localization
  And Mapping
}